\documentclass[conference]{IEEEtran}
\IEEEoverridecommandlockouts
\usepackage{cite}
\usepackage{amsmath,amssymb,amsfonts}
\usepackage{algorithmic}
\usepackage{graphicx}
\usepackage{textcomp}
\usepackage{xcolor}
\def\BibTeX{{\rm B\kern-.05em{\sc i\kern-.025em b}\kern-.08em
    T\kern-.1667em\lower.7ex\hbox{E}\kern-.125emX}}
\begin{document}

\title{Image Coding for Machines with Object Region Learning\\
}
\author{\IEEEauthorblockN{Takahiro Shindo, Taiju Watanabe, Kein Yamada, Hiroshi Watanabe}
\vspace{5pt}
\IEEEauthorblockA{\textit{Graduate School of Fundamental Science and Engineering, Waseda University}\\
Tokyo, Japan \\
taka\_s0265@ruri.waseda.jp, lvpurin@fuji.waseda.jp, stslm738.ymd@toki.waseda.jp, hiroshi.watanabe@waseda.jp}
}

\maketitle

\begin{abstract}
Compression technology is essential for efficient image transmission and storage. 
With the rapid advances in deep learning, images are beginning to be used for image recognition as well as for human vision.
For this reason, research has been conducted on image coding for image recognition, and this field is called Image Coding for Machines (ICM).
There are two main approaches in ICM: the ROI-based approach and the task-loss-based approach. 
The former approach has the problem of requiring an ROI-map as input in addition to the input image. 
The latter approach has the problems of difficulty in learning the task-loss, and lack of robustness because the specific image recognition model is used to compute the loss function.
To solve these problems, we propose an image compression model that learns object regions.
Our model does not require additional information as input, such as an ROI-map, and does not use task-loss. 
Therefore, it is possible to compress images for various image recognition models. 
In the experiments, we demonstrate the versatility of the proposed method by using three different image recognition models and three different datasets. 
In addition, we verify the effectiveness of our model by comparing it with previous methods.
\end{abstract}


\section{Introduction}
In recent years, images and videos have become ubiquitous in our lives.
People are using them for social network services to enrich their lives.
Image compression is an important technology for handling a lot of images and videos.
It is essential, especially on occasions where many images need to be transmitted and stored while having limited bandwidth and storage.
For this reason, image coding methods such as JPEG\cite{a1}, AVC/H.264\cite{a2}, HEVC/H.265\cite{a3}, and VVC/H.266\cite{a4} have been created.
These image compression methods are composed of hand-crafted algorithms, created based on the knowledge of data encoding experts.
Neural network based image compression (NIC) has also been the subject of much research in recent years.
Many NIC models beyond VVC, the latest image compression standard, have been proposed and are expected to be widely used in the future\cite{a5,a6}.
These technologies are image coding methods for human vision, and coding performance is evaluated in terms of bitrate and image quality.

Meanwhile, with the development of image recognition technology, opportunities to use techniques such as object detection\cite{a7,a8} and segmentation\cite{a9} are rapidly increasing.
Conventional coding methods compress images while preserving image quality.
In other words, they are not efficient compression methods for image recognition.
Hence, it is necessary to devise an image compression method specifically for object detection models and segmentation models.
The research field on image compression for such purposes is called Image Coding for Machines (ICM)\cite{a10}.
There are two main approaches in the study of ICM.
The first approach is ROI-based method in which ROI-map is used to allocate more bits to the object region in the images. 
As shown in Fig. 1(a), this approach needs an ROI-map as input in addition to the image to be encoded. 
The problem with this approach is that it requires process to prepare the ROI-map before compressing the image.
The second approach is task-loss-based method.
As shown in Fig. 1(b), in this approach, the NIC model is trained using task-loss to create an image compression model for image recognition\cite{a13,a14}.
The task-loss is calculated by the image recognition accuracy of the coded image created using the NIC model.
For example, when training an NIC model for object detection, the detection accuracy of the coded image by NIC model is used as the loss function\cite{a15}.
Thus, the loss function is defined by the output values of the image recognition model.
However, these values are output from a black box, which makes it difficult for the NIC model to learn them.
In addition, when training an NIC model with task-loss, the NIC model corresponding to the image recognition model is required. 
It is due to the variation in task-loss, dependent on the type of image recognition model.

To solve this problem, we propose an NIC model which learns the object region in the images.
This compression model is trained using a Object-MSE-loss.
The Object-MSE-loss is the difference between the object region in the input image  and that of the output decoded image.
By applying this loss to train the NIC model, only the object regions is decoded cleanly leaving the other regions untouched.
Thus, information in the image that is unnecessary for image recognition is eliminated.
In the experiments, we create an NIC model trained with the proposed loss function and compare it with the SOTA image coding model for human vision\cite{a6}.
It is also compared with the image coding model for machines\cite{a14} to demonstrate the effectiveness of the proposal.
Furthermore, by using multiple datasets and multiple image recognition models, we show that our model is a robust to changes in image recognition models and changes in datasets.
\begin{figure}[bt]
    \centerline{\includegraphics[width=1\columnwidth]{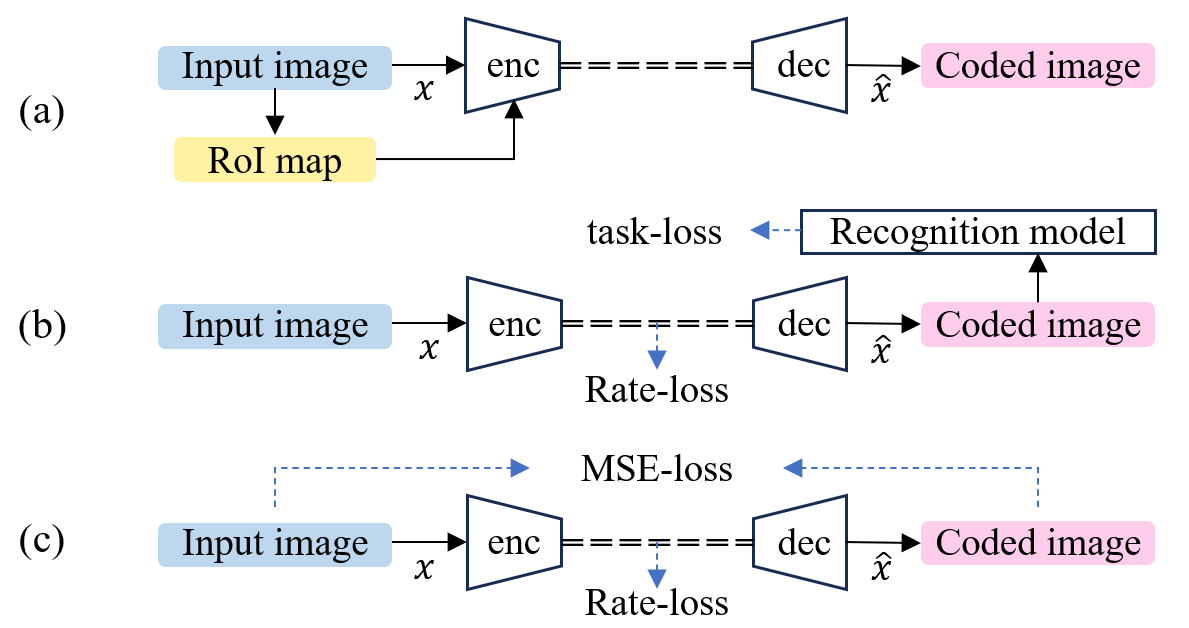}}
    \caption{Image compression process. (a) : NIC model for human vision. (b) : ROI-based approach for ICM. (c) : task-loss-based approach for ICM.}
    \label{fig:related}
    \end{figure}

\section{Related Works}
\subsection{Image Coding for human vision}
Image compression is an essential technique for efficiently transmitting and storing images. 
Research on these technologies has been ongoing for decades, and many image and video coding methods have been created. 
Examples of image and video coding standards include HEVC and VVC. 
VVC is the latest standard and is based on a hand-crafted algorithm. 
The first version of this compression method was standardized in 2020.
Video coding for an intra frame corresponds to image coding.

In recent years, there has been a lot of research on image compression using neural networks\cite{a16,a17,a18,a19}. 
G. Toderici \textit{et al}. proposed an image compression method using RNNs and created the first NIC model that exceeds the performance of JPEG\cite{a20}. 
Later, autoencoder-based methods emerged, including CNN-based NIC models and Transformer-based models\cite{a21,a22}. 
In addition, a mechanism called hyperprior was proposed to efficiently compress the features in the output of the encoder\cite{a23}. 
Hyperprior effectively captures spatial dependencies in the latent representation and is incorporated in many of NIC models.
These NIC models targeting human vision are trained using the following loss functions:
\begin{equation}
\mathcal{L}_{h}=\mathcal{R}(y)+\lambda \cdot MSE(x,\hat{x}).
\end{equation}
In (1), $y$ is the encoder output of the NIC model, $\mathcal{R}(y)$ is the bitrate of $y$ and is calculated using compressAI\cite{b1}. 
$x$ represents the input image, and $\hat{x}$ represents the decoder output image of the NIC model.
$MSE$ represents the mean squared error function and $\lambda$ is a constant to control the rate.
As shown in Fig. 1(c), the compression model is trained to restore the raw image while reducing the bitrate.

The latest NIC model is proposed by J. Liu \textit{et al}\cite{a6}. 
This model has a structure that combines CNN and Transformer.  
CNN is good at capturing features in a narrow region in the image, while the Transformer is good at capturing features in a wide region in the image\cite{a24}. 
By combining these advantages, this model has a coding performance that exceeds not only VVC but also other NIC models.

\subsection{Image Coding for Machines (ICM)}

As the performance of artificial intelligence using deep learning improves, opportunities for using image recognition techniques are increasing. 
The amount of image data used for this purpose is increasing,and image compression methods for machines need to be created.
ICM mainly considers image coding for image classification models\cite{a25}, object detection models\cite{a15}, and segmentation models\cite{a26}.

There are two main approaches in ICM. 
The first one is the ROI-based approach, which encodes images to allocate more bits to the object regions.
As shown in Fig. 1(a), the ROI-map is input to the encoder along with the image to be compressed.
When using image compression methods based on hand-crafted algorithms, neural networks are utilized to create ROI-map by predicting the location of objects in the image. 
H. Choi \textit{et al}. proposed an ICM method that combines an object detection model, YOLO9000\cite{a27}, and an image compression method, HEVC\cite{a11}.
The backbone of YOLO9000 extracts features from the image, and the likely location of the object is predicted from the features. 
Then, using HEVC's rate control system, image coding is performed to allocate more information to the object region.
Z. Huang \textit{et al}. proposed a method that uses a Region Proposal Network to predict the location of objects in an image and compresses the image using VVC based on the prediction results\cite{a12}. 
In addition, B. Li \textit{et al}. proposed an ROI-based method using the NIC model\cite{a28}. 
In this method, the input is an image for compression and its corresponding ROI map.
They showed that these image compression methods outperform basic HEVC and VVC in terms of image compression performance in image recognition accuracy.

The second approach in ICM uses task-loss to train the NIC model to code images for image recognition\cite{a29,a30,a31}. 
As shown in Fig. 1(b), the output encoded image of the NIC model is input to the image recognition model, and the task-loss is calculated from the result. 
When training the NIC model for object detection, the loss is calculated from the object detection accuracy of the coded image\cite{a15}, 
and when training the NIC model for image classification, the loss is calculated from the image classification accuracy of that\cite{a25}. 
However, image recognition model is black box, and training NIC models using task-loss is difficult. 
Therefore, in general, the NIC model is trained using a loss function that adds task-loss to (1). 
In this case, the loss function can be expressed as follows:
\begin{equation}
    \mathcal{L}_{m}=\mathcal{R}(y)+\lambda_{1} \cdot MSE(x,\hat{x})+\lambda_{2} \cdot \mathcal{M}(\hat{x}).
\end{equation}
In (2), $\mathcal{R}$, $MSE$, $y$, $x$, and $\hat{x}$ have the same meaning as those functions, variables, and constants in (1).
$\mathcal{M}(\hat{x})$ is the task-loss that can be computed by inputting the coded image into the image recognition model.
$\lambda_{1}$ and $\lambda_{2}$ are constants to control the rate.

A different method from these is the Omni-ICM model proposed by R. Feng \textit{et al}\cite{a14}. 
This model employs Information Filtering (IF) to remove information from an image, that is unnecessary for image recognition. 
IF module is trained to eliminate image redundancy while maintaining the feature obtained when the image is input to ResNet\cite{a32}. 
This method produces images with less entropy and maintains image recognition accuracy.
Since the model is not trained using a specific task-loss, it can compress images for many different image recognition models.

\begin{figure}[bt]
    \centerline{\includegraphics[width=1\columnwidth]{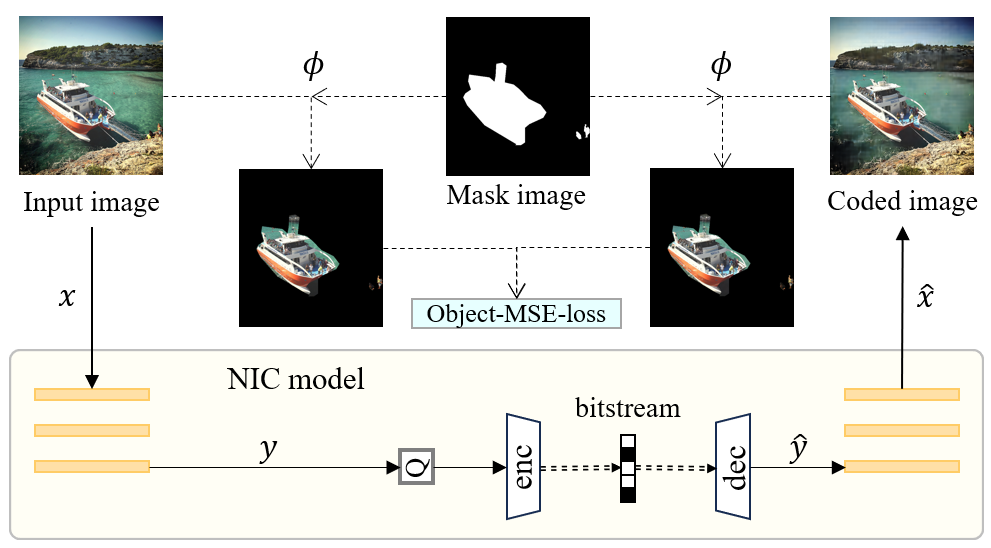}}
    \caption{The proposed training method of the NIC model.}
    \label{fig:loss}
    \end{figure}

\section{Proposed Method}
The problem with the ROI-based ICM approach is that the object region in an image must be predicted before image compression process. 
In this approach, an image compression model uses the ROI map obtained from object detection models or manual derivations. 
This compression method needs the ROI map to allocate more bits to object region in the image.
Therefore, the ROI-based ICM approach requires both the image and its ROI map as inputs.
The problem with the task-loss-based approach is that the NIC model is difficult to train. 
This is because the task-loss is calculated from the results obtained by inputting the coded image into the image recognition model, which makes the loss calculation complex.
Another point to consider is that if the NIC model is trained with only task-loss, image reconstruction is difficult. 
Therefore, they are often trained by adding MSE-loss to task-loss, which is inefficient as a compression model for image recognition.

Considering these problems, we propose an NIC model that learns the object region in the image. 
The Object-MSE-loss is used to train proposed NIC model. 
This loss represents the MSE-loss of the object region in the image. 
By training the NIC model with the Object-MSE-loss, we can create a compression method that only decodes the object region in the image. 
The proposed training method of the NIC model is shown in Fig. \ref{fig:loss}.
The mask image is a binary mask with object regions set to 1 and all other regions set to 0.
This binary mask is used to create an image in which all regions except the object region are black.
This image creation process is represented by the Hadamard product of the image and the corresponding binary mask as follows:
\begin{equation}
    \phi(a,m_{a})= a \odot m_{a}
\end{equation}
In (3), $\phi$ is a mask function that blackens non-object pixels of the image, and $a$ represents the input image.
$m_{a}$ is the binary mask corresponding to $a$.
Using this mask function, the Object-MSE-loss is calculated, which is expressed as follows: 
\begin{equation}
    Object\_MSE(b,c)=MSE(\phi(b,m_{b}),\phi(c,m_{c})).
\end{equation}
In (4), $b$ and $c$ represent certain images, and $\phi$ is as shown in (3).
$m_{b}$ and $m_{c}$ are the binary masks corresponding to $b$ and $c$ respectively.
In addition, using (4), the loss function used to train the proposed NIC model is expressed as follows: 
\begin{equation}
    \mathcal{L}_{p}=\mathcal{R}(y)+\lambda \cdot Object\_MSE(x,\hat{x}).
\end{equation}
In (5), $\mathcal{R}$, $y$, $x$, $\hat{x}$, and $\lambda$ have the same meaning as those functions, variables, and constants in (1).
The proposed NIC model only learns how to encode and decode the object region in the image. 
For the other regions of the image, it learns to reduce the $\mathcal{R}(y)$.
Due to the reduction of $\mathcal{R}(y)$, we achieve a rough texture surrounding the object in the image.
The proposed NIC model is effective as an image encoding method for image recognition because it can cleanly restore only the object region in the image.

\begin{figure*}[bt]
    \centerline{\includegraphics[width=2\columnwidth]{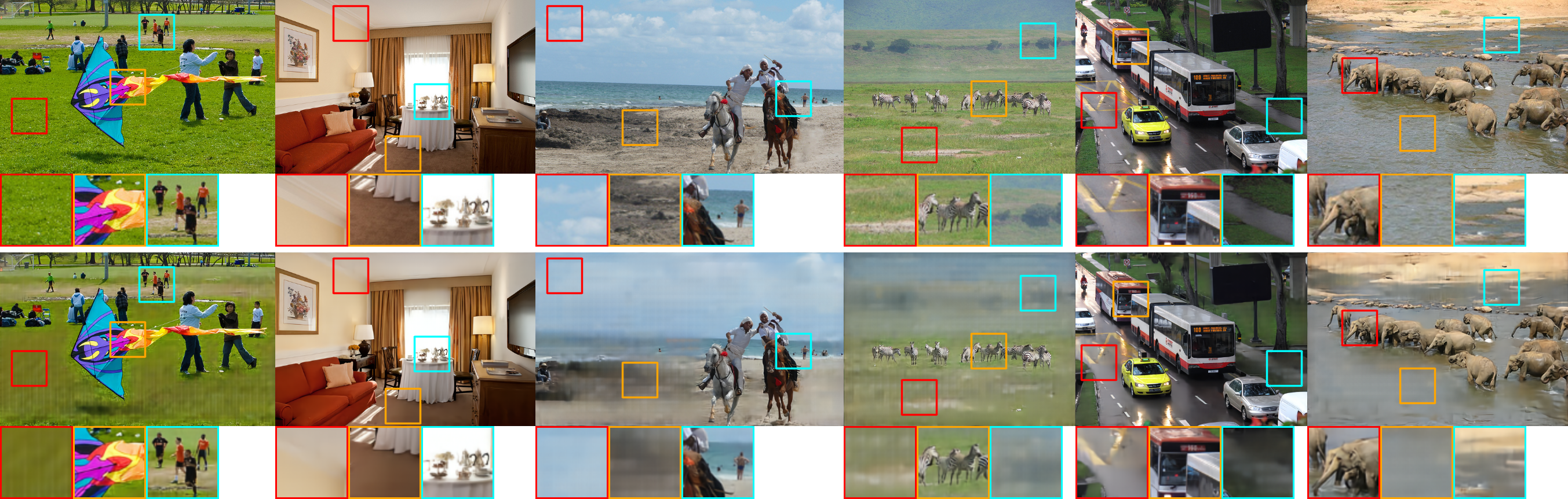}}
    \caption{Examples of coded images of the COCO2017 dataset with the proposed NIC model. The upper and lower rows are the input and output coded images, respectively.}
    \label{fig:exp}
    \end{figure*}

\section{Experiments}
\subsection{Training image compression model}
We modify the loss function used to train the NIC model proposed by J. Liu \textit{et al}\cite{a6}. 
This NIC model is originally an image compression model targeting human vision and is trained using (1) as the loss function. 
To change this model to an image compression model for machines, we train this model using (5) as the loss function. 
The dataset used for training is COCO2017\cite{a33}. 
This dataset is for instance segmentation and object detection, which consists of 118287 images for training (COCO-Train) and 5000 images for validation (COCO-Val).
The ground-truth segmentation map of this dataset is used to create a mask image for calculating the Object-MSE-loss.
Only the COCO-Train is used for training. 
In (5), four different $\lambda$ (0.05, 0.02, 0.01, 0.005) are used to create four NIC models.

As for a comparison, an NIC models trained with the loss function from (1) are prepared. 
To ensure a fair comparison, COCO-Train is used for training, and four types of $\lambda$ (0.01, 0.005, 0.002, and 0.001) are used to create four NIC models.

\subsection{Evaluation Method}
We measure the image compression performance of the created NIC models in terms of image recognition accuracy. 
YOLOv5\cite{a34}, Mask-RCNN\cite{a35}, and Panoptic-deeplab\cite{a36} are used as image recognition models. 
YOLOv5 is an object detection model, Mask-RCNN is an instance segmentation model, and Panoptic-deeplab is a panoptic segmentation model. 
The instance segmentation model can perform object detection simultaneously, while the panoptic segmentation model can perform instance segmentation at the same time.

First, we investigate the image compression performance in object detection accuracy of YOLOv5. 
We use COCO2017 and VisDrone\cite{a37} as the datasets for object detection. 
These datasets for validation are encoded with two models: the proposed model and the comparison model. 
Coded images produced by the proposed method are shown in Fig. \ref{fig:exp}.
The output images of the NIC model trained using Object-MSE-loss are clean in the object region and unclear in the non-object region.
To measure the detection accuracy, we use YOLOv5 trained on regular images for detecting objects within the coded images. 
Additionally, we encode the training dataset with the proposed model and the model for comparison. 
By applying these encoded datasets to YOLOv5, we obtain a fine-tuned YOLOv5. 
Furthermore, we measure the object detection accuracy of the fine-tuned YOLOv5 by letting it detect objects in coded images.

Next, we investigate the compression performance of Mask-RCNN in terms of object detection accuracy and instance segmentation accuracy.
We use MMDetection\cite{a38} to implement Mask-RCNN.
We encode the COCO-Val with the two models we have just created. 
The encoded images are input to Mask-RCNN trained with the original images to measure the image recognition accuracy. 
Furthermore, the COCO-Train is encoded to prepare Mask-RCNN fine-tuned with these images.
Encoded images for validation are also input to this fine-tuned model to measure the image recognition accuracy.

In addition, we examine image compression performance in terms of semantic segmentation and instance segmentation accuracy using Panoptic-deeplab.
Using the Cityscapes dataset\cite{a39}, the validation data and training data are coded with two models, the proposed model and the model for comparison. 
The coded images of the training dataset are used for fine-tuning the Panoptic-deeplab. 
The coded images for validation are input to the fine-tuned model, and the image recognition accuracy is measured.

\begin{figure*}[bt]
    \begin{minipage}[b]{0.48\linewidth}
        \centering{\includegraphics[width=1\columnwidth]{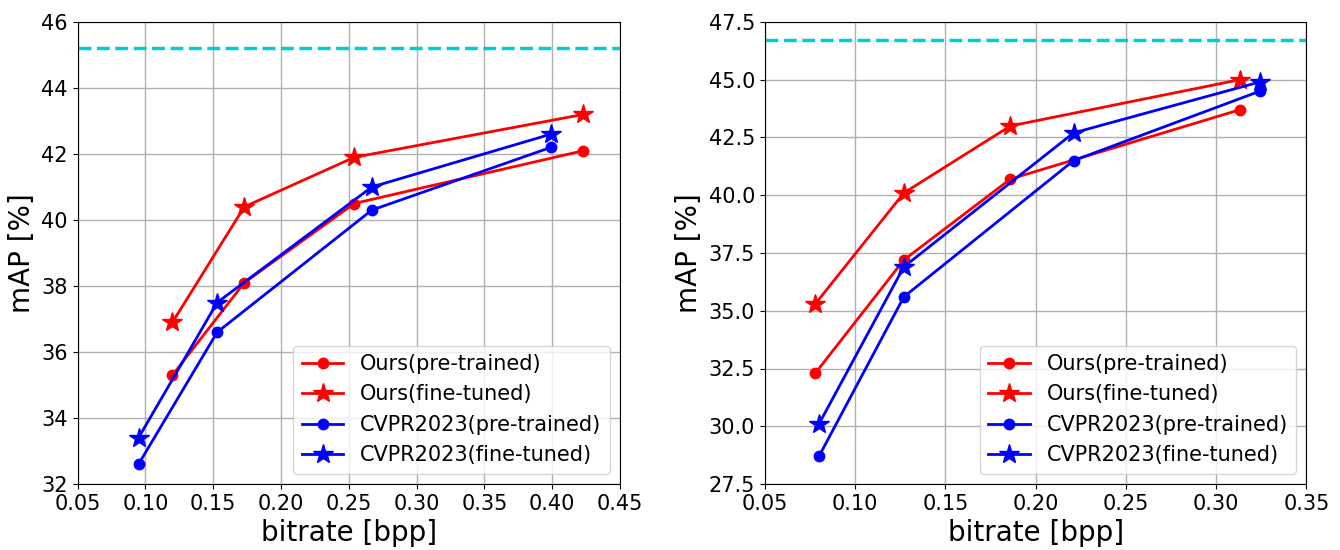}}
        \caption{Compression performance in object detection accuracy of YOLOv5. The left figure shows compression performance for COCO, and the right figure shows the same for VisDrone.}
        \label{fig:coco}
      \end{minipage}
    \begin{minipage}[b]{0.48\linewidth}
        \centering{\includegraphics[width=1\columnwidth]{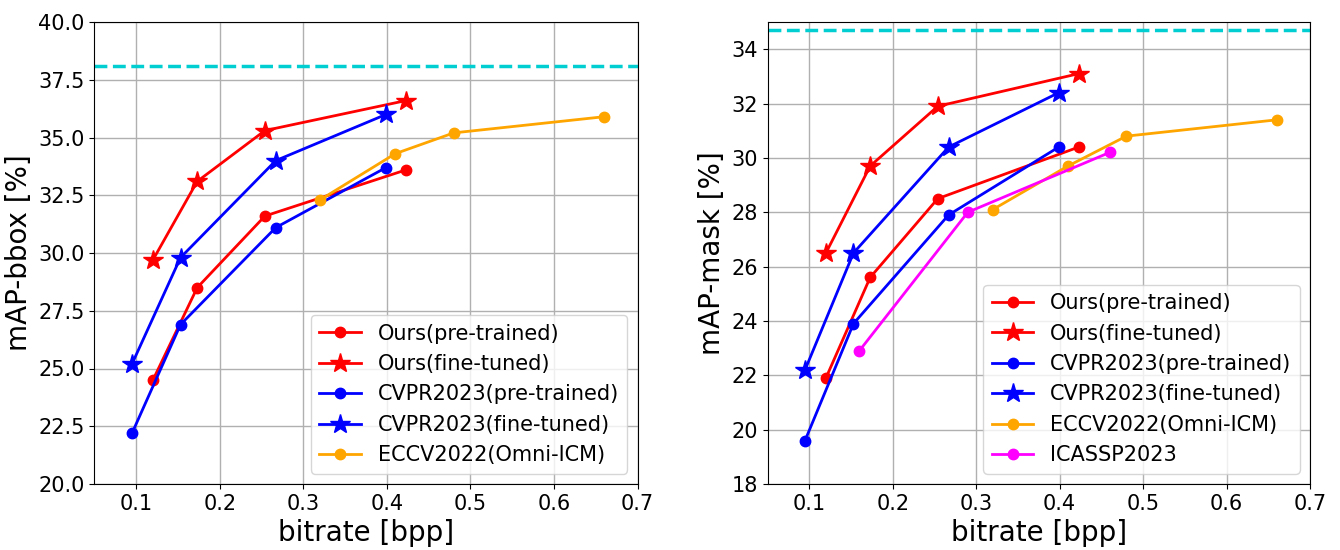}}
        \caption{Compression performance in image recognition accuracy of Mask-RCNN. The left and right figures show the compression performance in detection accuracy and instance segmentation accuracy, respectively.}
        \label{fig:mask}
      \end{minipage}
    \end{figure*}
\begin{figure*}[bt]
    \begin{minipage}[b]{0.48\linewidth}
        \centering{\includegraphics[width=1\columnwidth]{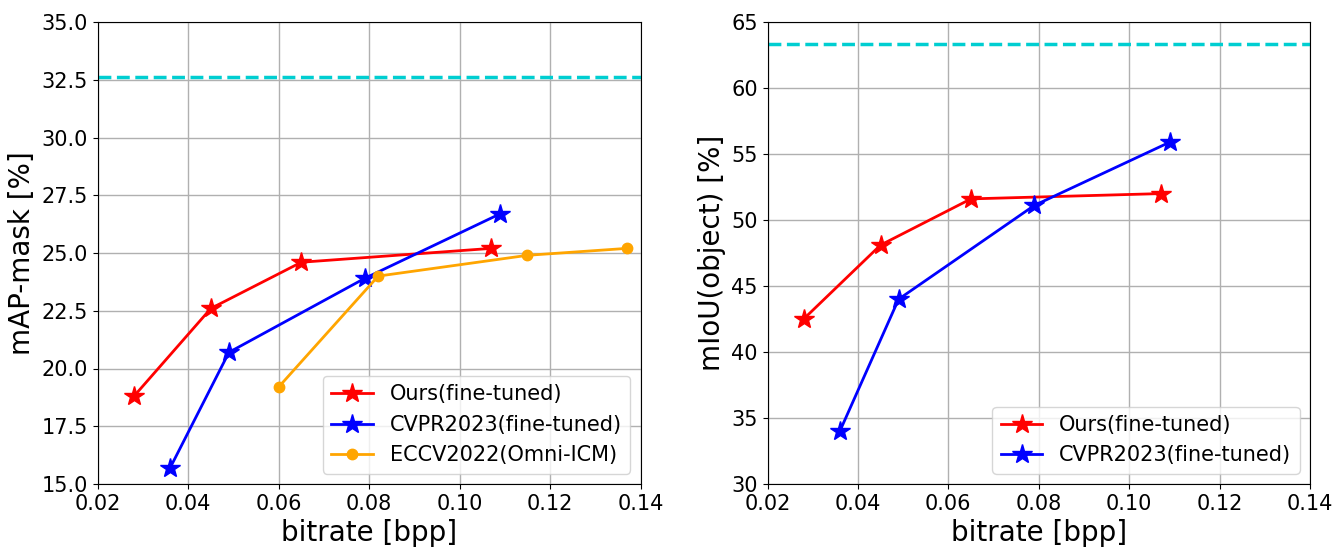}}
        \caption{Compression performance in instance segmentation accuracy of Panoptic-deeplab.}
        \label{fig:city}
        \end{minipage}
    \begin{minipage}[b]{0.48\linewidth}
        \centering{\includegraphics[width=1\columnwidth]{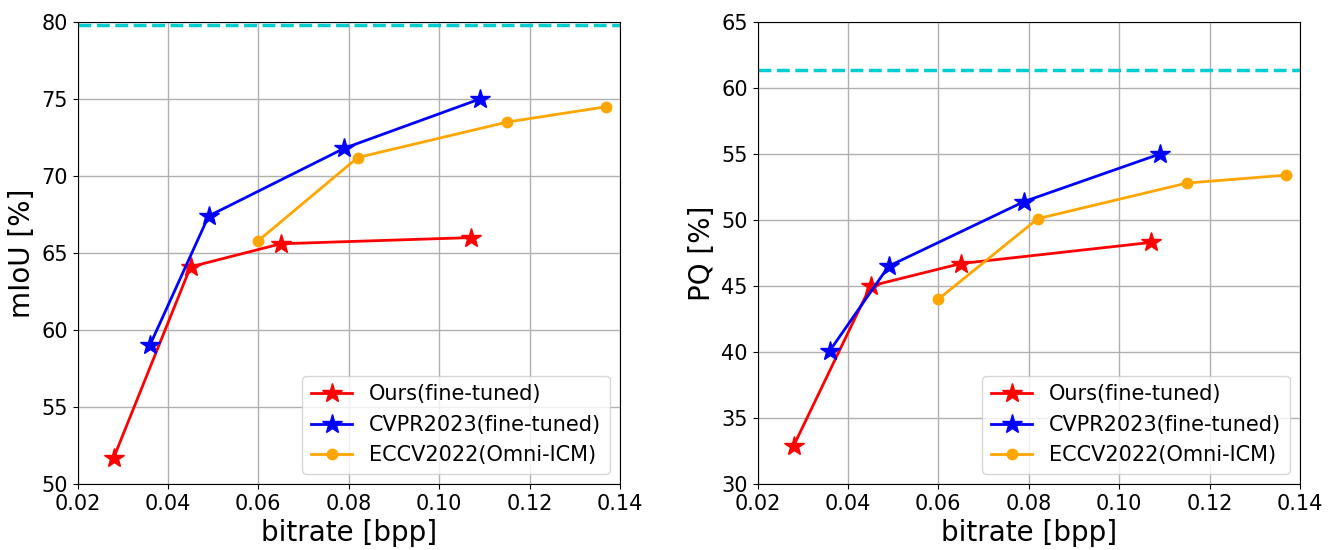}}
        \caption{Compression performance in panoptic segmentation accuracy of Panoptic-deeplab.}
        \label{fig:city1}
        \end{minipage}
    \end{figure*}
\begin{figure}[bt]
    \centerline{\includegraphics[width=1\columnwidth]{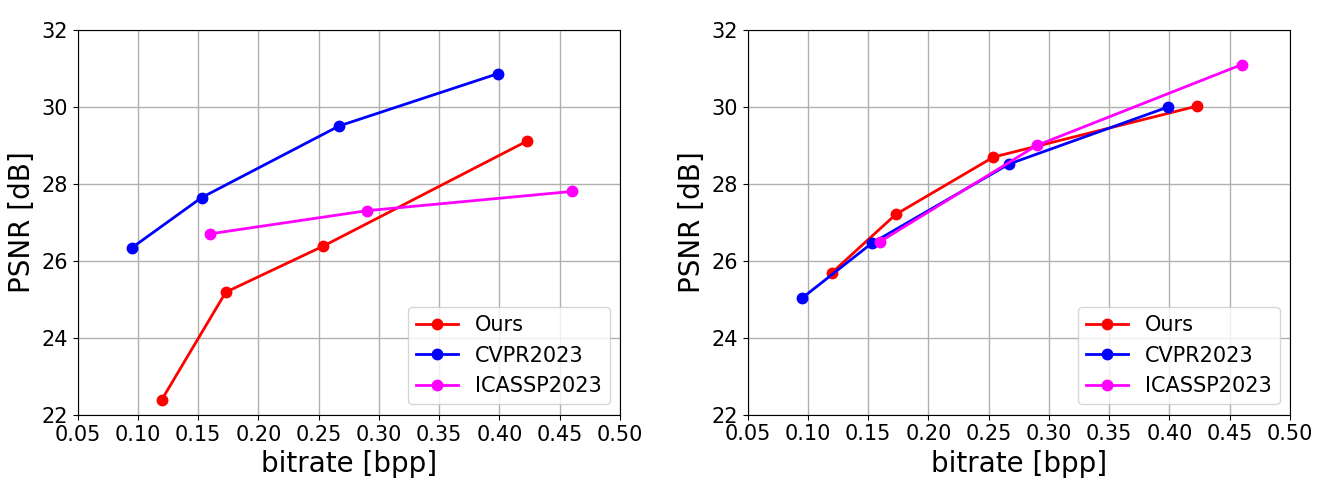}}
    \caption{Compression performance in image quality. The left figure shows the image quality of the entire image, and the right figure shows that of the object region in the image.}
    \label{fig:psnr}
    \end{figure}

\subsection{Results}
The results of object detection accuracy using YOLOv5 on coded images are shown in Fig. \ref{fig:coco}. 
The light blue dotted line indicates the image recognition accuracy of the uncompressed image.
The red line shows the coding performance of the proposed method, and the blue line shows that of the method for comparison. 
For both methods, the detection accuracy in YOLOv5 trained using the original images and the detection accuracy in YOLOv5 fine-tuned using the coded image are measured. 
In Fig. \ref{fig:coco}, the left graph shows the validation results using COCO dataset, and the right graph shows the validation results using VisDrone dataset.
Fig. \ref{fig:coco} shows that the best coding performance is obtained when the proposed compression method is used for the compression model and the fine-tuned YOLOv5 is used for the object detection model. 
Moreover, even when the original YOLOv5 is used, the proposed method outperforms the comparison method at low bitrates. 
These results are related to the $\lambda$ values in the loss function used when creating the compression model. 
Four values of $\lambda$, 0.05, 0.02, 0.01, and 0.005, are used when creating the proposed model, and 0.01, 0.005, 0.002, and 0.001 are used when creating the comparison model. 
Since the proposed method does not cleanly decode all regions of the image but the object region, the bitrate is unlikely to increase even if $\lambda$ is set to a large value. 
In other words, at the same bitrate, the proposed method may decode the object region in the image more cleanly than the comparison method. 
Fig. \ref{fig:psnr} shows the PSNR of the entire image and that of only the object region in the image in the coded image. 
It can be seen that the proposed method is inferior to the comparison method in the image quality of the whole image but outperforms in that of only the object region at low bitrates.

Furthermore, in Fig. \ref{fig:coco}, the proposed coding model shows a larger performance improvement by fine-tuning the object detection model compared to the conventional coding model. 
One of the reasons for this is that the method for comparison tries to recover the original image, while the proposed method does not. 
Since the comparison method tries to decode an image close to the original one, 
the difference in object detection accuracy between YOLOv5 trained on the original image and YOLOv5 trained on the coded image is small. 
On the other hand, the proposed model does not attempt to decode images close to the original images, 
so the YOLOv5 trained on the original image cannot achieve sufficient detection accuracy.
Therefore, fine-tuning the YOLOv5 with the coded image can greatly improve the coding performance.

The results of object detection and instance segmentation using Mask-RCNN on the coded image are shown in Fig. \ref{fig:mask}. 
The red and blue lines represent the same meaning as in Fig. \ref{fig:coco}.
The orange lines indicate the coding performance of the image compression model proposed by R. Feng \textit{et al}\cite{a14}.
The pink lines indicate the coding performance of the image compression model proposed by B. Li \textit{et al}\cite{a28}. 
In Fig. \ref{fig:mask}, the left graph shows the relationship between object detection accuracy and bitrate, while the right graph shows the relationship between instance segmentation accuracy and bitrate.
In both cases, the best coding performance is achieved by utilizing the proposed image coding model for image compression and Mask-RCNN fine-tuned with the coded images.

The results of instance segmentation and panoptic segmentation with Panoptic-deeplab on the coded image are shown in Fig. \ref{fig:city} and Fig. \ref{fig:city1}.
All red, blue, and orange lines represent the same meaning as in Fig. \ref{fig:mask}. 
Fig. \ref{fig:city} shows the results of instance segmentation. 
It can be seen that the proposed method is effective at low bitrates, outperforming both the existing and comparative methods. 
On the other hand, Fig. \ref{fig:city1} shows the results of panoptic segmentation, where the proposed method is ineffective. 
This is because panoptic segmentation does a segmentation for the object region and the non-object region in an image.
In the output image of the proposed compression model, regions other than object regions are blurred and unclear. 
Therefore, although the proposed method is effective for instance segmentation, it is unfit for panoptic segmentation.

\section{Conclusion}
We proposed an NIC model which learns the object region in images.
By training the NIC model with object-MSE-loss, we created a model that decodes object regions of images intensively. 
Experimental results show that the proposed NIC model is effective as an image encoding method for image recognition, especially for object detection and instance segmentation. 
The experiments used three different image recognition models and three different datasets to show that the proposed encoding model can be used robustly. 
Future work is required to further improve the encoding performance by reducing the texture of object region while maintaining the image recognition accuracy.

\section*{Acknowledgment}
These research results were obtained from the commissioned research (No.05101) by National Institute of Information and Communications Technology (NICT), Japan.

\vspace{12pt}

\end{document}